# Real-Time Knee Angle Prediction Using EMG and Kinematic Data with an Attention-Based CNN-LSTM Network and Transfer Learning Across Multiple Datasets

Mojtaba Mollahossein[1], Gholamreza Vossoughi*[2], Mohammad Hossein Rohban[3]

[1] Department of Mechanical Engineering, Sharif University of Technology, Tehran, Iran. Email: mojtaba.hoseini@mech.sharif.edu

[2*] Department of Mechanical Engineering, Sharif University of Technology, Tehran, Iran. Email: vossough@sharif.edu

[3] Department of Computer Engineering, Sharif University of Technology, Tehran, Iran. Email: rohban@sharif.edu

**Abstract**

Electromyography (EMG) signals are widely used for predicting body joint angles through machine learning (ML) and deep learning (DL) methods. However, these approaches often face challenges such as limited real-time applicability, non-representative test conditions, and the need for large datasets to achieve optimal performance. This paper presents a transfer-learning framework for knee joint angle prediction that requires only a few gait cycles from new subjects. Three datasets - Georgia Tech, the University of California Irvine (UCI), and the Sharif Mechatronic Lab Exoskeleton (SMLE) - containing four EMG channels relevant to knee motion were utilized. A lightweight attention-based CNN-LSTM model was developed and pre-trained on the Georgia Tech dataset, then transferred to the UCI and SMLE datasets. The proposed model achieved Normalized Mean Absolute Errors (NMAE) of 6.8 percent and 13.7 percent for one-step and 50-step predictions on abnormal subjects using EMG inputs alone. Incorporating historical knee angles reduced the NMAE to 3.1 percent and 3.5 percent for normal subjects, and to 2.8 percent and 7.5 percent for abnormal subjects. When further adapted to the SMLE exoskeleton with EMG, kinematic, and interaction force inputs, the model achieved 1.09 percent and 3.1 percent NMAE for one- and 50-step predictions, respectively. These results demonstrate robust performance and strong generalization for both short- and long-term rehabilitation scenarios.

## 1- Introduction

Electromyography (EMG) measures electrical signals generated by contracting muscle fibers, reflecting neuromuscular activity. [1]. EMG is typically measured using electrodes placed on the skin's surface (surface Electromyography (sEMG)). Alternatively, electrodes may be inserted into the muscle tissue [2]. The frequency range of EMG signals is generally reported to be from 6 to 500 Hz, with most power concentrated between 20 and 250 Hz [3]. Analyzing EMG signals provides valuable information about muscle activation patterns, coordination, and fatigue levels. This information makes EMG signals invaluable in biomechanics, rehabilitation, and prosthetics [4]. EMG analysis has become applicable in various fields, particularly in the development and control of lower limb exoskeletons. Using EMG, exoskeletons can provide real-time assistance to individuals with mobility impairments to enhance their walking gait through adaptive control mechanisms [5]. Assisting a subject with abnormality in the lower limb to complete the walking phase is one the most important aspects of exoskeleton control application in rehabilitation. The gait cycle consists of two primary phases: the stance and the swing phase [6]. The stance phase accounts for approximately 60% of the gait cycle, and the swing phase constitutes the remaining 40%. The knee joint is a vital element in these two primary walking phases. So, accurate prediction of knee angles using EMG signals could be a crucial step in real-time environments such as exoskeleton control, facilitating a more natural and efficient gait pattern for users [6].

The knee joint's movement is influenced by the coordinated activity of several key muscles, each of which contributes to a specific part of knee flexion and extension during the walking phase. The coordinated activity of muscles, including the Vastus Medialis (VM), Vastus Lateralis (VL), Rectus Femoris (RF), Biceps Femoris (BF), Semitendinosus, Semimembranosus, and Gastrocnemius, plays a critical role in knee joint movement during the gait cycle. Some other muscles also affect the knee joint, although their effect is more superficial than the muscles reviewed ([7], [8]). Although there are numerous advantages to using EMG signals, there are also significant limitations, including signal noise, crosstalk, variability across individuals, electrode placement inconsistency, muscle fatigue effects, and the complexity of data acquisition protocols ([9],[10]). These challenges necessitate advanced computational techniques, such as machine learning (ML) and deep learning (DL), to process EMG signals effectively and extract meaningful patterns [11]. Numerous machine learning (ML) methods have been applied to EMG signals for various classification and regression tasks ([12], [13], [14], [15], [16], [17]). While a few of these methods achieve acceptable precisions in some specific scenarios and protocols, they often face limitations that make their application in real-time experiments challenging, if not impractical. These limitations include:

1. **Preprocessing constraints**: Many techniques rely on preprocessing steps that are not applicable in real-time scenarios. For example, preprocessing steps such as low-pass and de-noising filters are often applied to the entire test of the subject, requiring all samples from a test to be available beforehand, which is impractical in real-time applications ([13], [17]).
2. **Cross-validation issues**: Cross-validation methods often inadvertently allow models to access validation or test data from the same subjects included in the training process, particularly when EMG signals are used in solidarity, leading to overfitting and reduced generalizability in real-world scenarios ([13], [17])
3. **High sampling frequencies**: Many studies process EMG signals at their original sampling frequencies (e.g., 1000 or 2000 Hz) ([17], [13], [14]). However, after some standard preprocessing, the effective frequency of the processed EMG signal is typically below 30 Hz. this is shown in the GitHub respiratory Therefore, down sampling the signal to 100 Hz does not lead to aliasing [19], as the effective frequency content of the preprocessed EMG signal remains well below the Nyquist frequency (50 Hz), ensuring signal integrity is preserved. Reducing the input frequency can significantly decrease the size of the input data, making it more efficient for network processing. Some datasets have also provided EMG signals at lower frequencies, such as 100 Hz, to facilitate processing and real-time applications [20].
4. **Network optimality**: Studies combining EMG and kinematic data as network inputs, usually report that the addition of EMG does not always lead to noticeable improvements [12]. This result is often obtained from suboptimal network designs for processing EMG signals. EMG is a complex signal, so the network must first be optimized on this signal before evaluating its synergistic effects with kinematic data. This prevents overfitting on the kinematic data.

The most effective way to evaluate deep learning methods is by comparing results on similar datasets or standardized benchmarks. Our primary objective is to predict knee angles for both normal and abnormal subjects in the UCI dataset [21]. As will be detailed in Section 3, this dataset contains a highly limited number of walking gait cycles per subject. However, its inclusion of subjects with knee abnormalities makes it highly valuable for rehabilitation analysis. Due to the scarcity of available data, most studies have focused on classification tasks. Classification requires less training data, compared to regression tasks, due to less complexity, while regression-based analyses remain largely unexplored. Various ML and DL methods have been applied to this dataset for three-class classification (walking, flexion, and extension ([17], [16], [22]) and binary classification (normal vs. abnormal) ([23], [24]). However, our focus is on predicting knee angle data as a regression task. Two key studies that have conducted regression tasks on the UCI dataset are the following:

MyoNet utilizes a Long-term Recurrent Convolutional Network (LRCN) with transfer learning for lower limb movement recognition and knee joint angle prediction from sEMG signals. The model achieved an average Mean Absolute Error (MAE) of 8.1% for healthy subjects and 9.2% for abnormal ones [13]. Multi-model fusion-based Ridge Regression (MMF-RR) has been introduced for knee joint angle prediction using EMG and historical joint angles [14]. The model integrates four machine learning algorithms and applies ridge regression with penalty terms to improve prediction accuracy. Evaluated on the UCI dataset, MMF-RR achieved an MAE of 6.73% for healthy participants and 8.97% for individuals with knee joint abnormality, outperforming several existing methods in motion intention recognition for rehabilitation applications. The UCI dataset provides valuable data from both normal and abnormal subjects; however, it lacks enough training data to effectively be trained in deep learning models. In contrast, some other datasets, which are accessible upon reasonable request, contain a significantly larger volume of data per subject, making them more suitable for deep-learning applications [20]. Although large-scale datasets are typically available for normal subjects, as is the case here, they often lack data for abnormal subjects. *The primary objective of this study is to leverage knowledge from a network trained on an extensive dataset and transfer it to a smaller dataset that contains valuable but limited information from rare abnormal subjects. This approach enhances the model's ability to generalize and improves prediction performance in real-world rehabilitation scenarios.*

In this study, we propose an attention-based CNN-LSTM network to predict real-time knee joint angle using EMG signals, both independently and in combination with kinematic data and the interaction forces (when the subject is with exoskeleton). By addressing challenges such as preprocessing constraints, cross-validation issues, and network optimization, this approach aims to perform real-time knee angle prediction according to EMG signals for new subjects utilizing transfer-learning. The network is first optimized using EMG signals alone, and then, kinematic and kinetic data are integrated to enhance prediction performance. The network's structure is designed to ensure low-volume network weights and reduce the size of the input data through carefully selected preprocessing steps. This lightweight design enables the network to be fine-tuned quickly and requires little data from the new subject, making it highly suitable for real-time applications.

**Technical Contributions:**
The main technical contributions of this study are as follows:

- Efficient signal processing: The EMG signals are downsampled to 100 Hz, enabling the design of a compact and computationally efficient network.
- Incorporation of attention mechanisms: An attention module is integrated into the network to enhance feature selection and improve predictive performance.
- Optimized lightweight architecture: The network structure is carefully designed to have a minimal number of parameters, balancing accuracy with computational efficiency.
- Data-efficient transfer learning: A transfer learning strategy is proposed that requires only a few gait cycles from new subjects, reducing the risk of overfitting and minimizing the need for extensive subject-specific data.
- Optimizing the network with EMG input before incorporating kinematic data: Before adding kinematic data as input, the network is first optimized using only EMG signals. This strategy mitigates the risk of overfitting kinematic data, ensuring that the network effectively learns from the EMG features.
- Selecting an EMG-based source dataset for pre-training: The source dataset is specifically chosen to include EMG signals, enabling meaningful transfer learning for the target task. In contrast, many existing studies adopt pre-trained models including convolutional neural networks, mostly used for image classification, which may not be optimal for EMG-based applications.
- Transfer learning was performed on datasets obtained from subjects walking both independently and while equipped with the exoskeleton under non-periodic gait conditions: When the subjects operated with the exoskeleton, only the robotic joint angles measured by the encoders in zero-force mode were available, rather than the true human knee angles. Under these circumstances, the interaction forces introduced additional disturbances and modulated the EMG signal amplitudes, making the knee angle prediction task substantially more challenging.

In summary, these contributions are combined to enable effective transfer learning with minimal data from new subjects, ensuring the method's applicability in real-world rehabilitation scenarios.

The remainder of this paper is organized as follows: The proposed methodology, including the network architecture, dataset's structure and transfer learning strategies, is described in Section 2. Experimental results are presented and analyzed in Section 3. Finally, conclusions and discussions are provided in Section 4.

## 2 Method

This section provides a detailed explanation of the methodology employed in this study, outlining the dataset, prediction scenarios, preprocessing steps, network architecture, and transfer-learning approach. The flowchart diagram of the Method Section is shown in Figure 1.

### 2-1 Dataset

In this study, three datasets were utilized: the open-source UCI dataset [21], the Georgia-tech dataset (obtained upon request) [20], and the SMLE dataset (acquired from our experimental laboratory tests). This dataset is not publicly available and will be provided upon request, with the citation of this paper for any use of the data. Our first target is the UCI dataset, which, despite having a limited number of samples per test, comprises both normal and abnormal tests. Accordingly, our strategy is to initially train the network on a comprehensive dataset (the Georgia-tech dataset) and subsequently transfer the trained model to the UCI dataset. Ultimately, the network will also be adapted to the experimental tests called SMLE dataset to further validate our network under conditions in which the subject is interacting with an exoskeleton.

### 2-1-1 Georgia-tech dataset

The Georgia-tech dataset is comprehensive, containing enough training samples, making it well-suited for the development of deep neural networks [20]. However, it lacks data from abnormal subjects, whose EMG signal analysis is crucial to detect walking phase abnormalities, an important key aspect of rehabilitation. In this dataset, 22 healthy subjects participated in performing different movements. In our study the subset of dataset, corresponding to treadmill walking, was employed. In this subset, subjects walked at 28 distinct speeds ranging from 0.5 m/s to 1.85 m/s, with increments of 0.05 m/s. These speeds were distributed across seven trials, with four speeds per trial, resulting in approximately 50 gait cycles per test. the data from four channels, analogous to the UCI dataset, were selected, as stated in Section 2-1-2.

**2-1-2 UCI dataset**

The UCI dataset includes both normal and abnormal subjects, making it valuable for rehabilitation-related studies. However, its primary limitation is the small amount of available data, which may hinder the network's ability to achieve acceptable precision [21]. A summary of the UCI dataset's characteristics is presented in Table 1.

Table 1- Information of UCI dataset [21]

| Aspect | Details |
| --- | --- |
| EMG Sampling Frequency | 1000 Hz |
| Number of subjects | 22 (11 healthy, 11 with knee abnormalities) |
| Channels | Biceps Femoris, Rectus Femoris, Semitendinosus, Vastus Medialis |
| Activities Recorded | Walking, Sitting with Knee Flexion, Standing with Knee Extension |
| Kinematic Data | Only knee angle measured by goniometer |

According to Table 1 in the UCI dataset, each subject performed three tests. In this study, only the data from walking test was employed. Also, the UCI dataset contains only four EMG channels. To ensure compatibility and enable transfer learning, only these four channels were considered from the Georgia-tech and the SMLE datasets described in Sections 2-1-1 and 2-1-3.

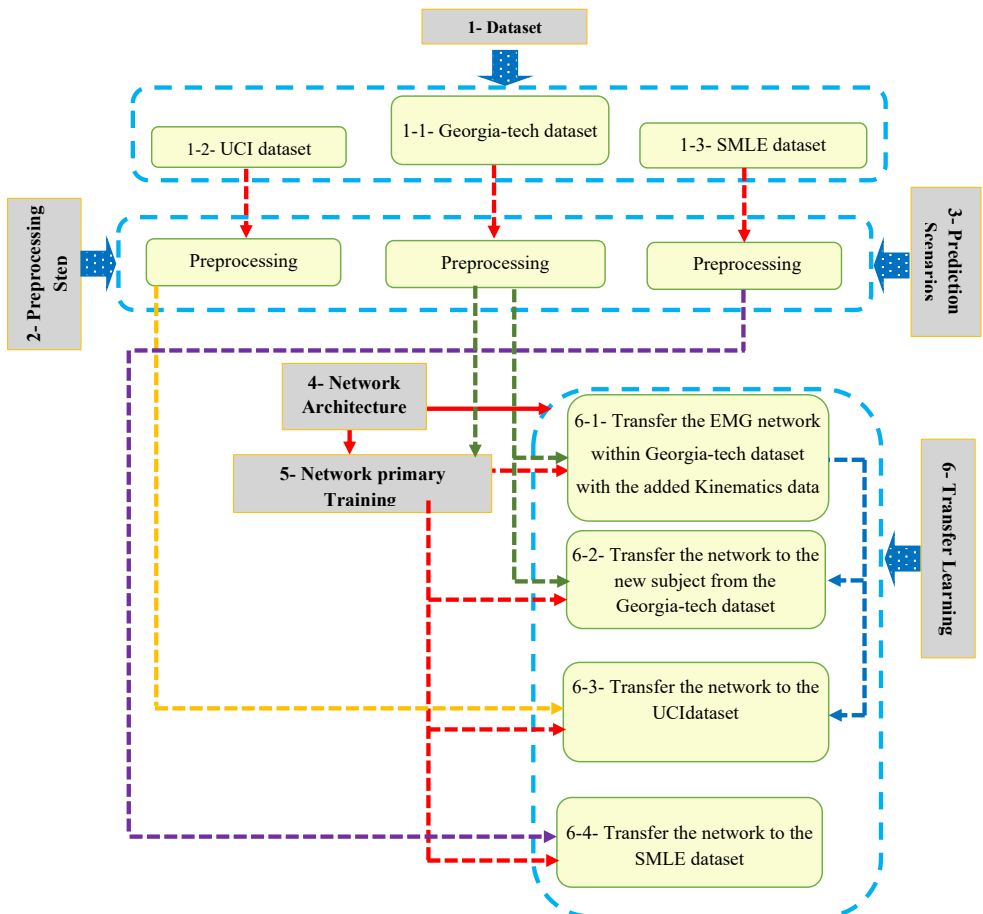

**Fig. 1-** A flowchart of the proposed methodology

### 2-1-3 Sharif-Mechatronic lab Exoskeleton (SMLE) dataset (Experimental test)

The data collection protocol in this section commenced with the placement of surface electromyography (EMG) electrodes on the right lower limb of a healthy subject. Electrodes were positioned over the muscle bellies of the Gluteus Maximus (GL), Vastus Lateralis (VL), Rectus Femoris (RF), Vastus Medialis (VM), Semitendinosus (ST), and Biceps Femoris (BF), and each channel was individually tested to ensure signal integrity. The subject was then carefully fitted with the lower-limb exoskeleton robot, with particular attention given to securing the straps without disturbing the underlying EMG electrodes. A zero-force controller was executed to minimize robotic impedance between the robot and the subject. In parallel, kinetic and kinematic data were collected, including thigh and shank interaction forces from embedded sensors and hip and knee joint angles from the exoskeleton encoders. For this trial, the subject's left leg was stabilized on the stationary edge of the treadmill, while the instrumented right leg executed a walking motion at a constant treadmill speed of 0.5 km/h. From this dataset, only the four EMG channels that aligned with the requirements specified in Section 2.1.2 were utilized, together with the knee joint angle and the thigh and shank interaction forces, all in 1000HZ frequency. The interaction forces were measured using calibrated load cells integrated into the exoskeleton. The shank load cell was positioned approximately 18 cm below the knee joint, and the thigh load cell was located about 30 cm below the hip joint

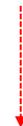

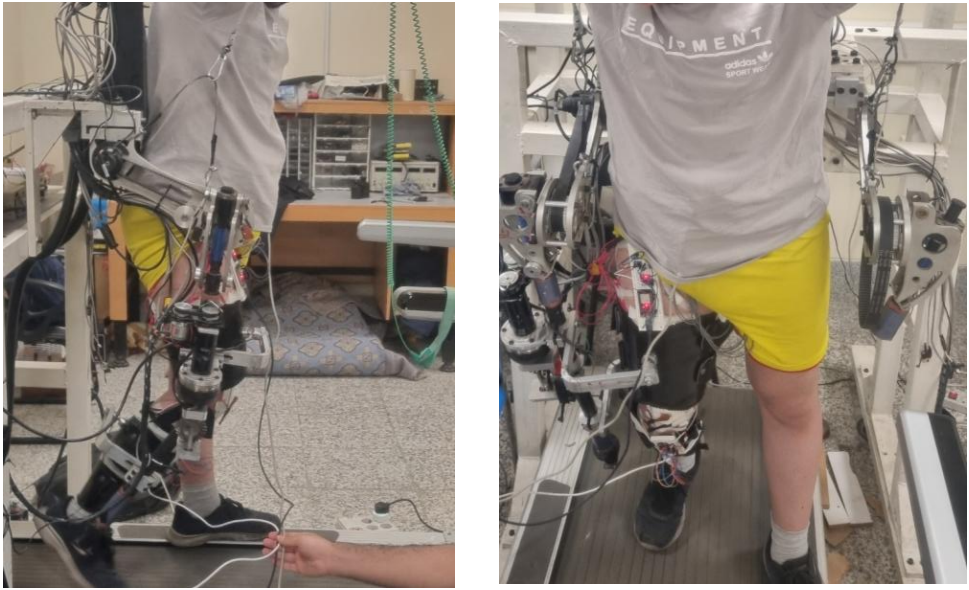

**Fig. 2-** Experimental test setup conducted in the Sharif-Mechatronic Lab

Figure 2 illustrates the human–robot interaction under zero-force mode control. During this experiment, IMU sensors were positioned on the shank and thigh; however, their data were excluded from the analysis. Only the EMG signals, the encoder-based knee and hip joint angles, and the shank and thigh interaction forces were retained for subsequent processing and analysis.

**2-2 Preprocessing steps**

Preprocessing is a crucial step in EMG analysis due to the complexity of the signal. Effective preprocessing can make the network more efficient by reducing input size and improving validation accuracy. A key aspect of this study is that all preprocessing steps are performed on an online window after segmentation, making the approach feasible for real-time applications. The preprocessing steps employed in this study are as follows:

1. High-pass filtering: A cutoff frequency of 20 Hz is applied to remove low-frequency noise and artifacts.
2. Recreation: All signals are converted to positive values by taking the absolute value of the filtered signal.
3. Low-pass filtering: A second-order Butterworth filter with a cutoff frequency of 3–6 Hz is applied to smooth the rectified signal.
4. Mean-variance normalization: The mean value is subtracted from the signal, and the result is divided by the standard deviation for each window.
5. Down-sampling: The sampling rate is reduced from 1000 Hz to 100 Hz to improve computational efficiency.

A window size of 2 seconds is used for all three datasets (with a sampling frequency of 1000 Hz for EMG data collection). A 2000-sample window at 1000 Hz is preprocessed and reduced to 200 samples at 100 Hz. These preprocessing steps are commonly employed in studies that predict lower-limb joint angles using EMG signals ([12], [20]).

**2-3 Prediction scenarios**

For all three datasets, four general scenarios were considered. These general scenarios involve predicting either a one-step or 50-step knee angle using (1) only EMG or (2) EMG in conjunction with knee angle data as inputs. Hereafter, we will refer to the first case as the single-input configuration (SIC) and the second case as the dual-input configuration (DIC). For the SMLE dataset, both the SIC and Dual-DIC scenarios were evaluated, with the interaction forces between the human and the exoskeleton (measured at the thigh and shank through embedded sensors) incorporated as additional inputs to the network. The details of splitting data into training, validation, and test sets were specified separately for each dataset in the following sections.

**2-4 Network architecture**

The proposed network architecture employs three key components. Convolutional Neural Networks (CNNs), Long Short-Term Memory (LSTM) networks, and the attention mechanism. CNNs are a kind of deep learning model designed to process data with a grid-like structure, particularly images and time-series data. They employ convolutional layers to automatically learn spatial features that effectively capture local patterns in the input data [25].

LSTMs are a type of Recurrent Neural Network (RNN) designed to model sequential data by learning long-term dependencies [26]. The attention mechanism enhances neural network performance by selectively focusing on the most relevant parts of the input sequence. This dynamic weighting improves the network's ability to capture important time and spatial information [27]. By integrating these components, the network is equipped with instruments to extract spatial features, model temporal dependencies, and dynamically prioritize critical input elements, resulting in superior predictive performance. The attention mechanism in our network is utilized only in DIC scenarios.

In the structure used for SIC in this research, four EMG channels from each window are fed to the network as the signal with a length of 200 (2s in 100Hz sample frequency). These four channels pass through the convolutional layers separately and are then concatenated. The concatenated layer is fed into two LSTM layers. After extracting time features with the LSTM layers, the last hidden state is fed to the final dense layer (containing 1 or 50 units according to the prediction horizon) to predict the knee angle. In DIC scenarios, the 200 kinematic data points are reshaped into a 50×4 matrix and fed into an LSTM layer, which processes the input across 50 consecutive time steps to extract temporal features. The network is designed such that the output of the CNN, before the concatenation layer, also produces 50 temporal features for each EMG channel. The attention mechanism operates between the hidden states of the LSTM, corresponding to the kinematic data, and the features extracted from the four EMG channels (before concatenation) across the 50-time segments. With this attention mechanism for each 50 time-steps, the model evaluates the significance of each channel to enhance feature extraction and improve prediction accuracy. The schematic of the network deployed in this research for DIC scenarios is shown in Figure. 3.

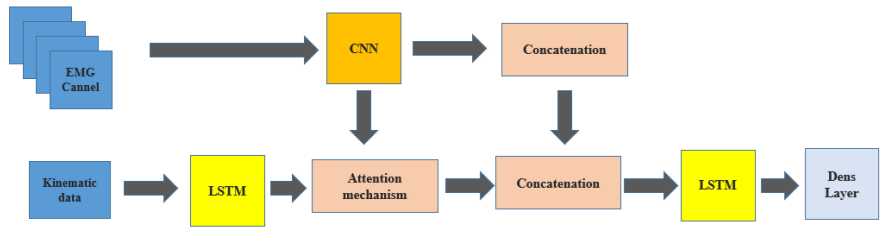

**Fig. 3-** The schematic of the proposed network (dual-input configuration (DIC))

Figure 3 illustrates the DIC scenario, while the remaining configurations are omitted for brevity and to maintain conciseness in the manuscript. In the SIC scenario, the upper part of Figure 3 is fed directly to the LSTM and Dense layer. In the scenario where the subject walks with the exoskeleton and interaction forces at the shank and thigh are measured, these values are fed through a convolutional network for processing. After the preprocessing stage, the extracted features are concatenated with the outputs of the final layers preceding the dense layer. Detailed implementation and further information of all scenarios are provided in the GitHub respiratory.

**2-5 Network primary training**

The partitioning methodology for the Georgia-tech dataset is designed to support real-time scenarios, particularly for new subjects (leave-one-out scenarios). The subjects of this dataset are divided into two groups: the first 21 subjects are allocated to the training and validation sets, while the remaining subject is considered as the test set. For the first 21 subjects, 80% of the data from each trial is used for training, and the remaining 20% is allocated for validation. This training is performed exclusively on the SIC scenario. We refer to this phase as the network's primary training, and the trained model is subsequently utilized in the following sessions for transfer learning. It is important to note that validation data partitioning is not applied during the test phase, and it is only employed to prevent overfitting during the training process.

**2-6 Transfer-learning**

The network trained from the Georgia-tech dataset is transferred to the new subjects from which the network has not seen any sample data. This means that the network is used with the weights from the previous training and will change with a lower learning rate compared to primary training to adapt to new data and use the history from past training in the network weights [29]. As discussed in Section 1, it is impossible to predict the joint angles of the new subjects in the SIC scenario without considering an adaptation phase on the new subjects. Due to unrepeatability of the EMG signals, the transfer-learning and fine-tuning of the network is mandatory. The transfer learning in this research is divided into 4 parts as follows:

**2-6-1 Transfer the EMG network within the Georgia-tech dataset with the added Kinematics data (SIC to DIC)**

The trained network from the Georgia-tech dataset in Section 2-5 is transferred to the DIC scenario in the same dataset. The network components that process the EMG signals remain unchanged, while the components handling the kinematic data are trained from scratch. It is important to note that in the DIC scenarios, if the network is simultaneously trained with the EMG and kinematic data, it will be biased toward the kinematic data, and the EMG part of the network will not be well optimized. The optimization of the EMG part of the network before adding the kinematic data is one of our method's merits that isolates this research from others.

**2-6-2 Transfer the network to the new subject from the Georgia-tech dataset**

In this section, transfer-learning is used for a new subject (test subject) from the Georgia-tech dataset. Each trial in the new test set is divided into fine-tuning and evaluation sets with a ratio of 0.5 to ensure effective model adaptation and assessment. For the SIC scenarios, the network described in Section 2-5 is transferred to the new subject. In the DIC scenarios, the pre-trained network in Section 2-6-1 is utilized. As illustrated in Figure 1, the flowchart includes two arrows originating from Sections 2.5 and 2.6.1, representing their corresponding connections in the subsequent scenarios. The network is fine-tuned using the fine-tuning data and evaluated using the evaluation data.

**2-6-3 Transfer the network to the UCI dataset**

Since this study focuses on predicting knee joint angle data during walking, only the walking class was considered in the UCI dataset. This results in 11 tests for normal subjects and 11 tests for abnormal subjects (one test per subject). All four general scenarios outlined in Section 2-3 were implemented to predict knee angles under two specific conditions: (1) fine-tuning exclusively on data from normal subjects and (2) fine-tuning exclusively on data from abnormal subjects. In each scenario, tests from 10 subjects were allocated to the training and validation sets, with a separation ratio of 0.8 for training and 0.2 in each test for the validation sets. The remaining test subject ($11^{th}$) was considered as the test set. Furthermore, the test set was divided into fine-tuning and evaluation subsets, with a split ratio of 0.5.

First, the network primarily trained with the Georgia-tech dataset, is re-trained with the training set and validated on the validation set to adapt to the new conditions (training on the population). The first training is referred to as first-stage training. Then, the network from the first stage is transferred to the new subject in the second-stage training with fine-tuning and evaluation sets. The pre-trained network from Section 2-5 is transferred for SIC scenarios, and the one in Section 2-6-1 is utilized for DIC scenarios.

**2-6-4 Transfer the network to the SMLE dataset**

In addition to EMG and kinematic data, shank and thigh force sensors were also available in the (SMLE) dataset as described in section 2-1-3. Four scenarios were considered: SIC, DIC, SIC-Forces, and DIC-Forces. In the SIC-Forces and DIC-Forces scenarios, the corresponding SIC and DIC setups were extended by including only the shank and thigh forces as additional inputs to the network. One healthy subject performed seven different tests. Six tests were used with an 80/20 data split in a shuffled manner for training and validation, allowing the network to adapt to the new experimental conditions. The final test was split evenly, with 50% of the data used for fine-tuning and 50% for evaluation. Accordingly, the scenarios corresponding to the first- and second-stage training, as described in Section 2.5.3, were considered in this analysis. The pre-trained network from Section 2.5 was transferred in all four scenarios, where the EMG portion of the network was fine-tuned with a lower learning rate, while the kinematic and force components were trained from scratch. So, as shown in Figure 1, the flowchart contains a single arrow originating from Section 2.5, with no corresponding connection from Section 2.6.1.

## 3 Results

This section presents the results corresponding to each scenario outlined in Section 2. The primary objective of this study is to achieve acceptable prediction accuracy on target subjects (new subjects in the UCI dataset and experimental data collected in the SMLE dataset). To maintain brevity and clarity in the main text, the complete training and testing results for Sections 2.5, 2.6.2, 2.6.3, and 2.6.4 are presented in the supplementary tables in the GitHub respiratory ,while Section 3.3 includes only the second-stage evaluation results for Sections 2.6.3 and 2.6.4.

### 3-1 Simulation parameters

As discussed in Section 2, all EMG data undergoes preprocessing and is subsequently down sampled to 100 Hz. A 2000 ms (millisecond) window with 200 samples is then fed into the network, with an overlap of 40 ms (4 samples) between consecutive windows. The network contains 72,361 parameters in the SIC scenario and 80,208 parameters the DIC one. For the SMLE dataset, the total number of trainable parameters amounted to 77,469 in the SIC-Forces scenario and 90,069 in the DIC-Forces scenario, respectively. Thus, the network is significantly lighter compared to deep learning architectures which typically contain parameters in the order of millions. In the primary training the batch size was 2000 (in transfer-learning scenarios, according to the number of training samples, a smaller batch size is considered), and the number of epochs is set between 40 and 80, depending on the scenario. Training is terminated if validation loss does not decrease after 5 consecutive epochs. The Adam optimizer is utilized with an initial learning rate of 0.001 for primary training (Sections 2-5 and 2-6-1). For transfer learning scenarios, the learning rate is reduced to either 0.0001 (1/10 of the original) or 0.0005 (1/5 of the original), depending on the specific scenario. By gradually reducing the learning rate, the network preserves essential features from the pre-trained model while allowing adaptation to the new data.

### 3-2 Performance evaluation criteria

To evaluate the network's performance across the scenarios outlined in Section 2, the metrics considered are Normalized Root Mean Squared Error (NRMSE), Normalized Mean Absolute Error (NMAE), and the Coefficient of Determination ($R^2$). These metrics are widely adopted in related studies for assessing EMG-based kinematic predictions ([12], [13],[14]).

### 3-3 Results for transfer-learning

In this section, the results for the target dataset (UCI dataset) in second-stage transfer-learning and the experimental SMLE dataset will be shown in the two following sections.

#### 3-3-1 Source and Target Data Ratio

It is crucial to understand the distribution of data across source and target datasets for a clearer interpretation of the results. The primary dataset, Georgia-tech, consists of data from 22 subjects, each evaluated at 28 different speeds, comprising over 120 gait cycles per subject. This vast dataset provides a substantial foundation for the initial training phase. In contrast, the UCI dataset includes 11 subjects, with abnormal data containing between 6 to 40 gait cycles per test, and the normal data ranging from 4 to 6 cycles. The SMLE dataset consists of 7 tests, with each test having between 4 and 15 gait cycles.

After preprocessing, the training and test samples were defined as follows: Georgia-tech includes 414,446 samples for training and validation, and 19,209 samples for testing, which incorporates both fine-tuning and evaluation with a 50% split ratio. For the UCI dataset, the normal subjects have 1,135 samples for training and validation, with 87 samples for testing. The abnormal subjects in the UCI dataset have 5,460 samples for training and validation, and 479 samples for testing. Finally, the SMLE dataset consists of 12,811 samples for training and validation, and 1,890 samples for testing

#### 3-3-2 Results for transferring the network to the UCI dataset

As mentioned in Section 2-6, the first stage transfer-learning is to adapt the network to the new environment. This adaptation (fine-tuning) is performed with 10 subject tests (training and validation sets). The fine-tuned network is transferred again to the new subject in the second stage (fine-tuning and evaluation sets). The results of the first stage fine-tuning on the UCI dataset are presented in supplementary tables in the GitHub respiratory. For the second fine-tuning on the new normal subject, the results corresponding to one-step and 50-step predictions, are presented in Tables 2 and 3, respectively. Similarly, for the new abnormal subject, the results for one-step and 50-step predictions are provided in Tables 4 and 5, respectively.

**Table 2-** Second network transfer to UCI dataset (normal subject and one-step prediction)

| Scenario | Evaluation NRMSE | Evaluation NMAE | Evaluation $R^2$ |
|---|---|---|---|
| SIC* | 0.100 | 0.068 | 0.864 |
| DIC** | 0.034 | 0.031 | 0.988 |

**Table 3-** Second network transfer to UCI dataset (normal subject and 50-step prediction)

| Scenario | Evaluation NRMSE | Evaluation NMAE | Evaluation $R^2$ |
|---|---|---|---|
| SIC* | 0.177 | 0.137 | 0.80 |
| DIC** | 0.044 | 0.035 | 0.988 |

**Table 4-** Second network transfer to UCI dataset (abnormal subject and one-step prediction)

| Scenario | Evaluation NRMSE | Evaluation NMAE | Evaluation $R^2$ |
|---|---|---|---|
| SIC* | 0.103 | 0.072 | 0.813 |
| DIC** | 0.030 | 0.028 | 0.984 |

**Table 5-** Second network transfer to UCI dataset (abnormal subject and 50-step prediction)

| Scenario | Evaluation NRMSE | Evaluation NMAE | Evaluation $R^2$ |
|---|---|---|---|
| SIC*....... | 0.189 | 0.156 | 0.697 |
| DIC**..... | 0.113 | 0.075 | 0.801 |

Tables 2 to 5 demonstrate the improvement in loss and error in the DIC scenarios rather than SIC ones. The effect of adding kinematic data is more pronounced in normal subjects compared to abnormal cases. This discrepancy may be attributed to the less periodic nature of abnormal gait cycles and noting that kinematic data provides greater predictive benefits in more periodic gait patterns. These results also indicate that EMG signals play a crucial role in predicting knee angles for abnormal subjects, where gait irregularities reduce the impact of kinematic inputs alone. Overall, the results indicate that predictions for normal subjects are generally more accurate than for abnormal subjects. This trend persists even when more training data is available for abnormal subjects in the UCI dataset; suggesting that knee angle prediction for abnormal subjects presents a greater challenge due to the increased variability and irregularities in pathological gait patterns. Additionally, the $R^2$ criterion in Tables 3 and 5, show that EMG alone does not provide acceptable prediction accuracy in the 50-step prediction scenario.

A comparison between Tables 2 to 5 with Tables related to the first adaptation to population in UCI dataset (detailed in supplementary tables in the GitHub respiratory reveals that first-stage transfer learning (using 10 subjects) achieves better performance than second-stage transfer learning (using one subject data with limited fine-tuning data from only three walking gait cycles for normal subjects and seven for abnormal ones). Figure 4 presents the results of knee angle prediction using EMG signals in normal test cases in a one-step prediction scenario.

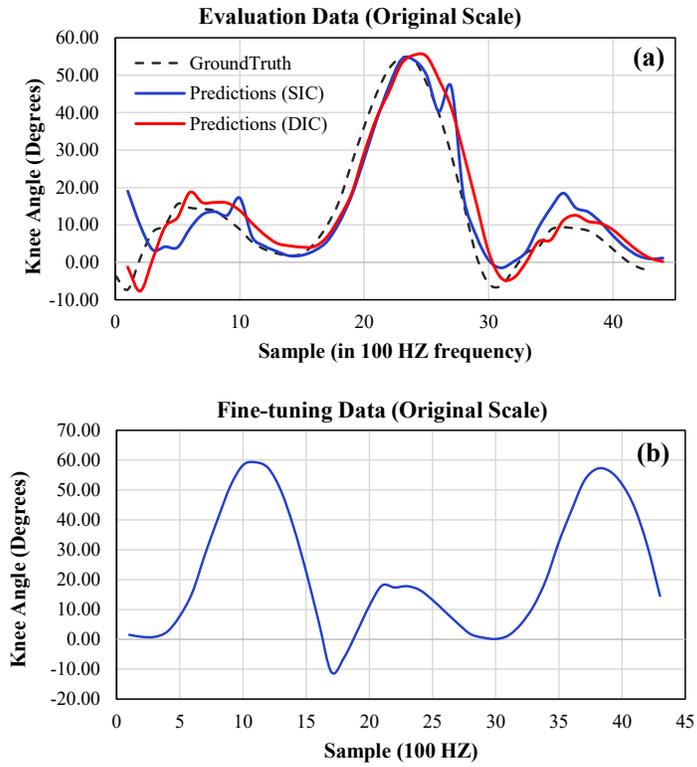

**Fig. 4-** The results of knee angle prediction using EMG signals in normal test cases in a one-step prediction scenario (UCI dataset)
a) Fine-tuning and evaluation data, b) evaluation data

Figure 4-a represents the fine-tuning data, while Figure 4-b illustrates the real and predicted values for the evaluation data in both SIC and DIC cases. It should be noted that the figures may not fully resemble a complete walking gait cycle due to the selection of an overlap exceeding one step, which is necessary for real-time scenarios. Figure 5 shows the results of knee angle prediction using EMG signals for abnormal test cases in a one-step prediction scenario.

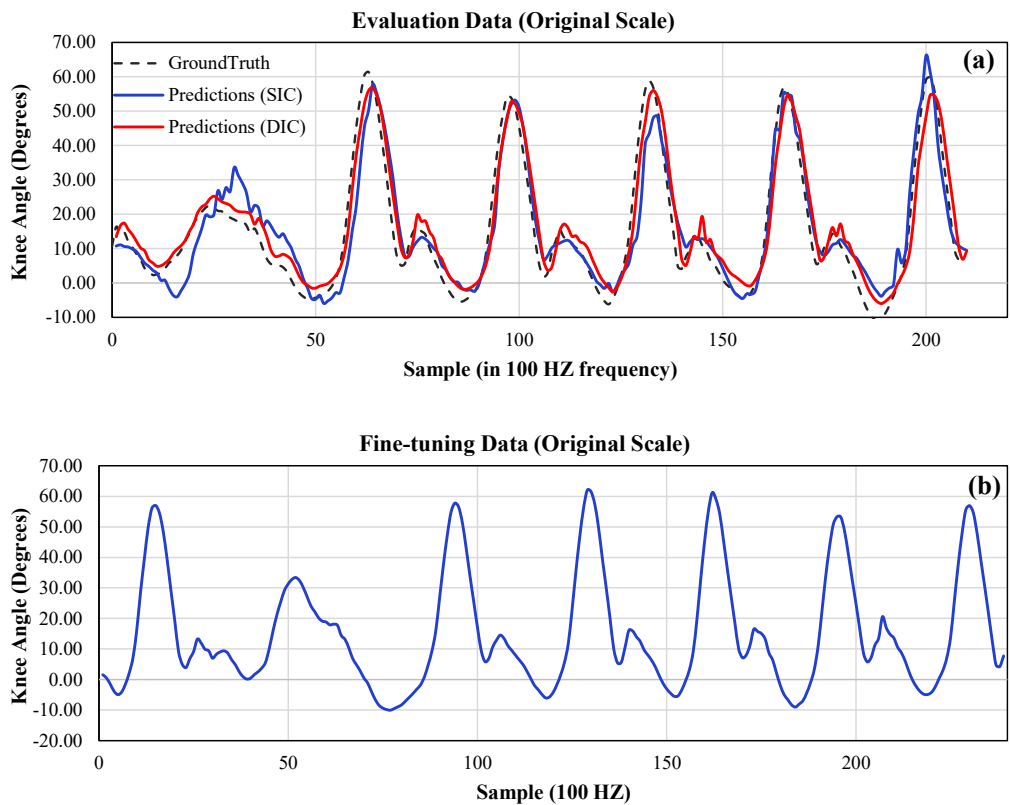

**Fig. 5-** The results of knee angle prediction using EMG signals in abnormal test cases in a one-step prediction scenario (UCI dataset) a) Fine-tuning and evaluation data, b) evaluation data

Figure 5-a represents the fine-tuning data, while Figure 5-b illustrates the real and predicted values for the evaluation data in both SIC and DIC cases. As shown in Figures 4-a and 5-a, the network demonstrates the ability to predict knee angle data using only EMG signals with very limited fine-tuning data (approximately three walking gait cycles for normal subjects and seven for abnormal subjects). These results highlight the model's capability to be effectively transferred to new subjects across different datasets, reinforcing its adaptability for real-world applications. Additionally, as observed in Figure 5, the irregular and non-periodic nature of gait cycles in abnormal subjects is evident.

At the end of this section, the last subject (out of 11) was replaced with a different subject to evaluate whether the network's transfer learning performance remains consistent across different new subjects. This experiment evaluated the robustness of the model, ensuring that its effectiveness is not highly sensitive to the specific new subject introduced during fine-tuning. The results of this part are also added in supplementary tables in the GitHub respiratory. The results show that the NMAE of the new subjects in different scenarios does not change more than 2 % in normal and abnormal subjects. It is worth emphasizing that the UCI dataset used in the transfer-learning stage constitutes only a minor portion of the Georgia Tech dataset employed during initial training. Specifically, the training data for the normal subjects account for about 0.27 %, while those for the abnormal subjects account for approximately 1.32 % of the total Georgia Tech training samples as stated in Section 3-3-1.

**3-3-3 Comparing the results with previous work**

As reviewed in Section 2, very few studies have conducted regression tasks on the UCI dataset. In this section, the results are compared with two primary regression studies on the UCI dataset ([13],[14]). Studies [13] and [14] implemented the SIC and DIC scenarios, respectively, for predicting knee joint angles from EMG signals. Accordingly, the corresponding scenarios were adopted in this work to enable a direct and consistent comparison with their methodologies. However, a direct comparison with these studies is not straightforward due to differences in methodology and evaluation approaches. The MyoNet study[13] performed regression using a CNN-LSTM network in the SIC scenario, employing a K-fold cross-validation method, whereas our work utilizes a leave-one-out validation approach that is more applicable in real-time scenarios. Additionally, in MyoNet, a 256 ms prediction window was used, while in our research, predictions were made for 10 ms and 500 ms windows. To compare our results with the Myonet we also consider a 26-step prediction (260 ms) to be more similar to the Myonet. The entire network in the SIC scenario from Section 2.5 is used, except for the final dense layer. A new dense layer with 26 units replaces the original dense layer with 1 unit. Subsequently, transfer learning is applied to adapt the network to the new configuration. To enable a more realistic and direct comparison with MyoNet, an additional evaluation was conducted using a 3-fold cross-validation strategy, aligning with the validation methodology adopted in MyoNet. This experiment specifically targeted the 260 ms prediction horizon, allowing for a consistent and meaningful comparison of performance under similar conditions. Table 6 presents a comparative NMAE analysis between our work and the MyoNet, highlighting key differences in methodology and performance.

Table 6- A comparative NMAE analysis between our work and Myonet (SIC* scenario) [13], [14]

| Methodology | Our method (leave-one-out scenario**) | Our method (3-Fold cross validation***) | Myonet |
|---|---|---|---|
| Normal subjects | 7.4 % in 260 ms prediction | 3.4 % in 260 ms prediction | 8.1 % in 256 ms prediction |
| Abnormal subjects | 9.29% in 260 ms prediction | 2.7 % in 260 ms prediction | 9.2 % in 256 ms prediction |

* Single-Input Configuration
** Considering new subjects for the test
*** Selecting 3 random data from the whole dataset as validation data

As shown in Table 6, our method achieves lower error rates compared to MyoNet, even under the more challenging leave-one-out validation scenario, except for the abnormal cases at the 260 ms prediction horizon. Notably, under the 3-fold cross-validation setting, our model significantly outperforms MyoNet, underscoring the effectiveness and optimality of the proposed architecture for knee angle prediction. Furthermore, as emphasized in this section, our study uniquely conducts evaluation on new, unseen subjects, representing a key advancement in the SIC scenario. As shown in Table 6, the results obtained under the 3-fold cross-validation scenario are notably better than those from the leave-one-out validation. However, it is important to note that while 3-fold cross-validation offers higher accuracy, it is less representative of real-world deployment, where the model must generalize to completely unseen subjects, as evaluated in the leave-one-out scenario.

The study reviewed in Section 2 [14] conducted the DIC scenario on the UCI dataset. Similar to MyoNet, it did not evaluate the model on new subjects. Instead, it utilizes an 80-20 training-validation split for model training in a random shuffling manner. Furthermore, according to the study, preprocessing was performed before segmentation, which is not suitable for real-time applications. Additionally, the model predicted knee angles at 1000 Hz (1 ms prediction intervals), whereas in our study, the shortest prediction horizon was one step ahead (equivalent to a 10 ms prediction interval).To allow a more meaningful comparison with MMF-RR [14], we designed an additional evaluation scenario in which the model predicts 10 ms ahead, and the validation sets are selected using a 3-fold cross-validation approach, consistent with the methodology used in [14]. Table 7 presents a comparative NMAE analysis between our approach and the average performance across different subjects from this study, which employed the Multi-Model Fusion-based Ridge Regression (MMF-RR) method (outperformed by several of their different ML and DL methods [14]

Table 7- A comparative NMAE analysis between our work and MMF-RR (DIC* scenario) [14]

| Methodogy | Our method(leave-one-out**) | Our method (3-Fold cross validation***) | MMF-RR |
|---|---|---|---|
| Normal subjects | 3.1 % in 10 ms prediction | 1.5% in 10 ms prediction | 6.58 % in 1ms prediction |
| Abnormal subjects | 3.5% in 10 ms prediction | 1.4% in 10 ms prediction | 9.18% in 1 ms prediction |

* Double-Input Configuration
** Considering new subjects for the test
*** Selecting 3 random data from the whole dataset as validation data

As shown in Table 7, the NMAE of our approach surpasses that of MMF-RR, even when predicting 10 ms knee angles and evaluating new subjects, highlighting a significant achievement in this study. Once again, the results under the 3-fold cross-validation setting exhibit improved performance; however, this evaluation strategy is generally considered less reliable for real-world applications compared to leave-one-out validation.

### 3-3-4 Results for transferring the network to the SMLE dataset

It is important to note that the actual human knee angles were not directly available from the dataset. As mentioned in Section 2-6-4 and only the knee joint encoder data from the exoskeleton was accessible. Consequently, the angles provided to the network as inputs and outputs in the different scenarios correspond to the robot's measurements rather than the subject's true joint kinematics. In the SML dataset, the interaction forces at the thigh and shank were also measured. The corresponding kinematic and kinetic profiles for Test 7 (the test set) are illustrated in Figure 6.

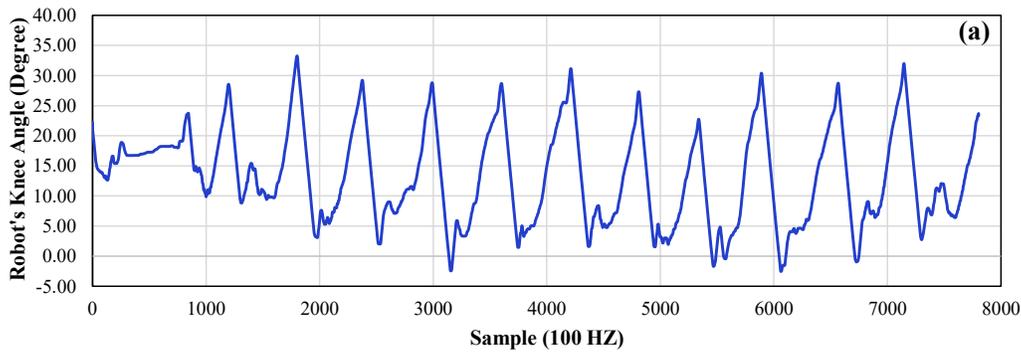

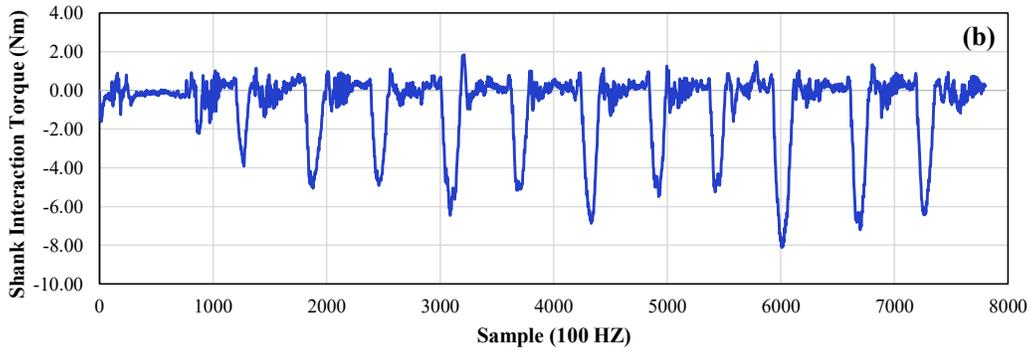

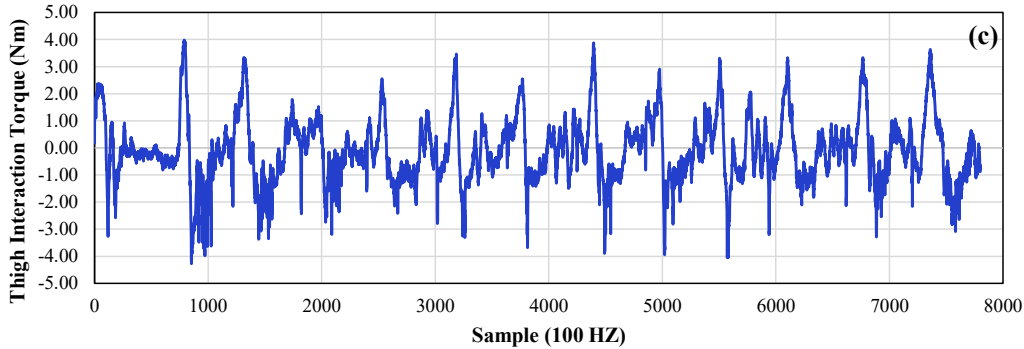

**Fig. 6-** the profile of kinematic and kinetic data sample in SMLE dataset a) Robot's knee angle, b) Shank interaction torque, c) thigh interaction torque

As illustrated in Figure 6, the kinetic data (shank and thigh interaction torques) are synchronized with the kinematic data (robot knee joint encoder angles). To enhance the interpretability of the dataset, Figures 6-b and 6-c depicts the corresponding joint torques at the shank and thigh segments rather than the raw interaction forces. The torques were obtained by multiplying each measured force by its respective moment arm distance from the corresponding robot joint. The data collected in the SMLE dataset were originally sampled at 1000 Hz; however, to ensure consistency with the preprocessed data used for network training, all signals were resampled to 100 Hz, and Figure 6 illustrates the results at this normalized frequency. As described in Section 2.6.5, the DML dataset is divided into training, validation, and test subsets as UCI dataset in Section 2-6-3. It is worth emphasizing that the SMLE dataset used in the transfer-learning stage represents only a small fraction of the Georgia Tech dataset utilized during the initial training phase. As stated in Section 3-3-1, the SMLE dataset contains approximately 3.1 % of the total source-domain data. The network was trained in two stages: initially on the training data, followed by fine-tuning on the designated fine-tuning subset. The outcomes for the four scenarios SIC, SIC-Forces, DIC, and DIC-Forces in the second stage of transfer-learning are presented in Tables 8 and 9, respectively, and include only the evaluation results. The results of the first-stage training and the fine-tuning data of the second stage are provided in the supplementary tables in the GitHub respiratory.

**Table 8-** Network transfer to the SMLE dataset (one-step prediction)

| Scenario | Evaluation NRMSE | Evaluation NMAE | Evaluation $R^2$ |
|---|---|---|---|
| SIC* | 0.1392 | 0.1101 | 0.6728 |
| SIC + Interaction Forces | 0.0976 | 0.0758 | 0.8393 |
| DIC** | 0.0171 | 0.0126 | 0.9951 |
| DIC + Interaction Forces | 0.0142 | 0.0109 | 0.9966 |

**Table 9-** Network transfer to the SMLE dataset ((50-step prediction))

| Scenario | Evaluation NRMSE | Evaluation NMAE | Evaluation $R^2$ |
|---|---|---|---|
| SIC | 0.1413 | 0.1112 | 0.6678 |
| SIC + Interaction Forces | 0.0976 | 0.0770 | 0.8414 |
| DIC | 0.0631 | 0.0468 | 0.9338 |
| DIC + Interaction Forces | 0.0461 | 0.0321 | 0.9647 |

As observed from Tables 8 and 9, the SIC and even the SIC-Forces scenarios were unable to accurately predict the knee angle. In both the one-step and 50-step prediction settings, the inclusion of interaction Forces and kinematic data substantially improved the prediction accuracy. The best performance was achieved in the DIC-Forces scenario, which incorporates both kinematic and kinetic information

The results of the one-step prediction on the test set, which includes both fine-tuning and evaluation data, across the four scenarios, are presented in Figure 7.

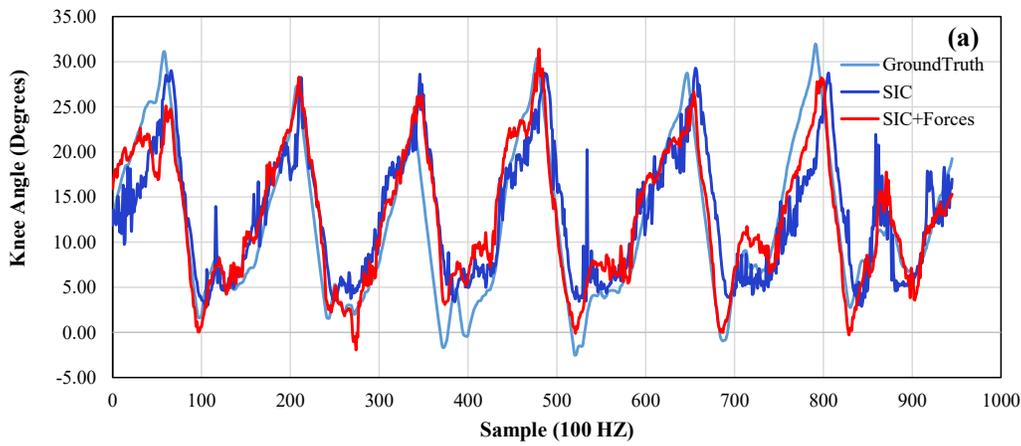

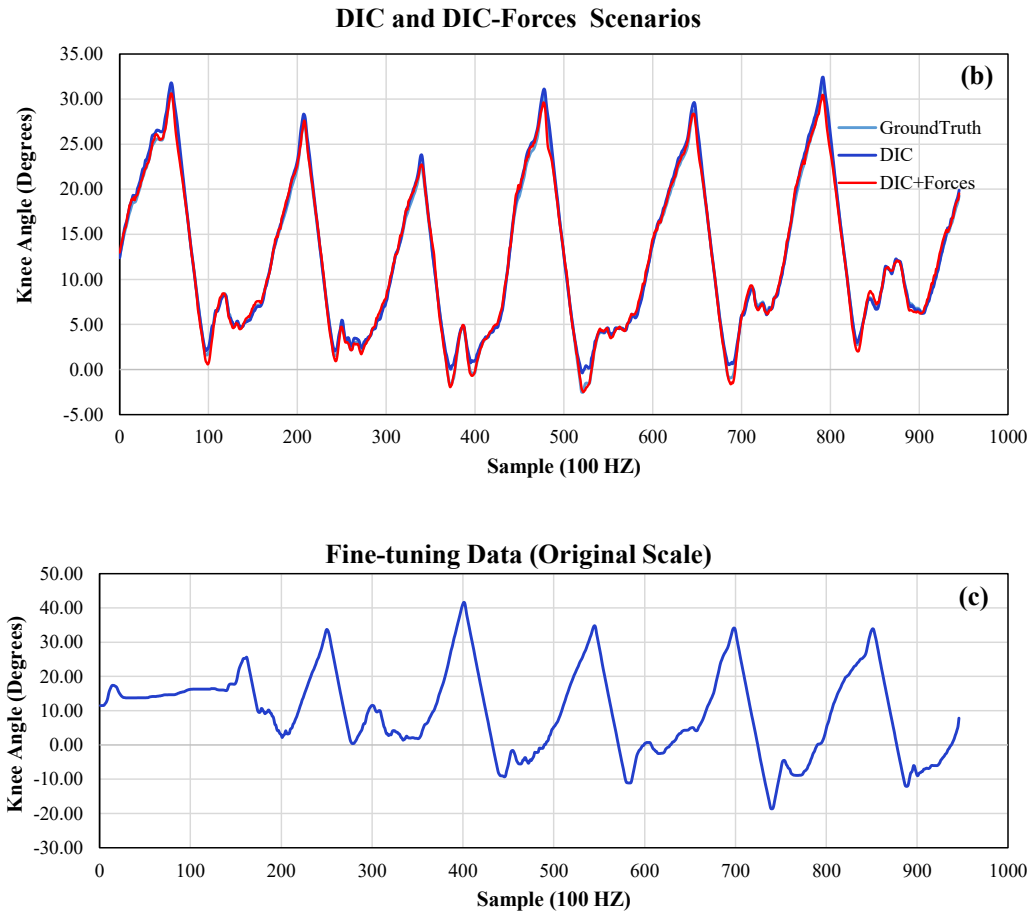

**Fig. 7-** Results for the new abnormal subject in a one-step prediction (SMLE dataset)
a) evaluation data (SIC and SIC+Forces), b) evaluation data (DIC and DIC+Forces), c) Fine-tuning data

As shown in Figure 7-a, the results for the SIC and SIC-Forces scenarios are not satisfactory, indicating that kinematic data are essential as network inputs to achieve accurate knee angle prediction. Moreover, in Figure 7-a and 7-b, it can be observed that incorporating kinetic (interaction) forces allows the network to better capture the peak regions within the gait cycle. The Figure 7-c indicate that, when the subject interacts with an exoskeleton, the DIC-Forces configuration provides the most robust and reliable performance by integrating all available input modalities.

**4-5 Concluding Remarks on the Results**

The results across the three datasets (Georgia Tech, UCI, and SMLE) provide valuable insights into the effectiveness of the proposed transfer learning approach for knee angle prediction using EMG signals. In the Georgia Tech dataset, where the model is trained on a large population dataset, the one-step prediction achieves high accuracy, even in SIC scenarios, suggesting that when the network is trained on extensive EMG data, it can generalize well within the same population in Supplementary tables (GitHub respiratory. The 50-step prediction, however, shows a noticeable decline in performance, highlighting the challenges of long-term forecasting. When tested on new subjects, the 50-step prediction remains unreliable, indicating the need for kinematic data and additional training data for long-term predictions. The incorporation of kinematic data improves prediction accuracy in both scenarios.

For the UCI dataset, the model demonstrates acceptable performance when fine-tuned on new subjects, with normal subjects achieving better results than abnormal ones. The addition of kinematic data further improves accuracy, particularly in normal gait patterns, where the gate cycle patterns are more periodic. However, for abnormal gait cycles, the impact of kinematic data is reduced due to higher variability and irregularities. The 50-step prediction with EMG alone is found to be unreliable, which could be attributed to the physiological delay in EMG activation (50–250 ms before muscle movement) and limited fine-tuning data in abnormal cases. Furthermore, the comparison of first-stage transfer learning (10 subjects) with second-stage transfer learning (only one subject with three gait cycles for the normal subject and seven for the abnormal subject) confirms that more training data is essential for long-term prediction accuracy. In Section 3-3-2, a comparative analysis was conducted between our work and two previous studies that performed regression tasks on the UCI dataset (one of our target datasets). The results demonstrate our achievement in improving prediction accuracy, even in more challenging and practical real-world scenarios, highlighting the applicability of our approach for rehabilitation.

In the SMLE dataset, like the UCI dataset, the first-stage training was conducted following the same transfer-learning framework. The model demonstrated strong performance, particularly in the DIC- Forces scenario for one-step prediction, highlighting the effectiveness of the transfer-learning approach in accurately predicting the robot knee encoder angle—even when the subject was interacting with the exoskeleton—rather than the direct human knee angle.

In summary, across all datasets, the one-step prediction consistently outperforms the 50-step prediction, demonstrating that short-term forecasting is more feasible with EMG-based models. The inclusion of kinematic data provides further improvements, particularly in cases with higher gait periodicity (normal cases). The results also confirm that transfer learning is effective, allowing the model to adapt to new subjects with minimal fine-tuning data, making it highly applicable for real-time rehabilitation scenarios without the need for extensive data collection. In both target datasets, there exists an important consideration that underscores the significance of the transfer-learning approach. In the UCI dataset, the inclusion of abnormal subjects introduces non-periodic and highly variable gait cycles, causing the prediction task to be inherently more complex. In the SMLE dataset, the data consists of robot encoder angles, where the presence of interaction forces between the human and the exoskeleton further increases the challenge of achieving accurate knee angle prediction. This configuration is particularly valuable for rehabilitation applications, where the estimation of joint angles is based solely on robotic encoder data, without relying on external measurement systems such as marker-based motion capture cameras or inertial measurement units (IMUs).

**5 Conclusion**

This paper presents a transfer learning-based approach for predicting knee joint angles using EMG data, with the goal of facilitating rehabilitation applications. The target of the paper is to leverage information from an extensive dataset (Georgia-Tech), which is typically available for normal subjects, to enhance predictive performance on a clinically valuable dataset (UCI dataset) containing rare abnormal subject data with knee pathologies. To address the challenge of limited abnormal subject data, the proposed method eliminates the need for data augmentation, which is often not appropriate for abnormal cases due to the uniqueness and variability of pathological conditions. The proposed transfer-learning framework was also validated on robot-assisted laboratory experiments conducted using a lower-limb exoskeleton, demonstrating its robustness in practical scenarios. This highlights the method's applicability to real-time rehabilitation environments, where marker-based motion capture systems, IMUs, and other external sensors may not be available for direct measurement of human joint angles

An attention-based CNN- LSTM network was initially trained using only EMG data from the Georgia Tech dataset (source dataset). Subsequently, the transfer learning was applied within the same dataset, integrating knee angle data. This strategy avoids overfitting kinematic inputs, ensuring the network maintains its generalization, which is one of the key distinctions of this study compared to previous works. Then, the transfer-learning approach was conducted on the target datasets (UCI and SMLE dataset) in two stages: Adaptation to the new conditions and fine-tuning to new subjects (in SMLE performed on a new test from the same subject), where the fine-tuning data consisted of only 3 to 7 gait cycles per test, which is practical in real scenarios. For the UCI dataset, the model achieved an NMAE of 6.8% and 13.7% for abnormal subjects in one-step and 50-step predictions, respectively. While for normal subjects, the corresponding NMAE values were 7.2% and 15.6%, respectively. These results were achieved using only EMG as the network input (SIC scenario), distinguishing this study by demonstrating exceptional predictive accuracy with minimal fine-tuning data. In the SMLE dataset, where complex human-robot interaction dynamics are present, accurate knee-angle prediction was not attainable under the SIC configuration. incorporating of the DIC configuration and interaction forces from the shank and thigh significantly enhanced the model's predictive capability, achieving an NMAE of 1.09 % for one-step prediction and 3.1 % for 50-step prediction, thereby demonstrating the effectiveness of the proposed transfer-learning framework under realistic exoskeleton conditions involving dynamic force exchange between the human and the robot.

The overall findings across both datasets highlight the proposed network's strong potential for real-time deployment in assistive and rehabilitation scenarios. Further investigations will focus on extending the transfer-learning framework to abnormal subjects performing walking tasks with exoskeleton.

**Data and code availability**

This study involved three datasets: one publicly available, one accessible upon request, and one locally collected. The UCI lower-limb EMG dataset is publicly available and does not require ethical approval. The Georgia Tech biomechanics dataset can be obtained directly from the research group via email request. The locally collected dataset and the source codes used for model implementation are available from the corresponding author upon reasonable request.

**Ethics approval**

a single human participant was involved in gait experiments conducted at Sharif University of Technology, Iran, using a lower-limb exoskeleton. All safety procedures for human–robot interaction were strictly followed, and the study protocol was reviewed and approved by the Biomedical Sciences Faculty Ethics Committee in Iran. The participant was fully informed about the experimental procedures and provided written consent prior to participation.

**References**

[1]  J. R. Cram, G. S. Kasman, and J. Holtz, "Introduction to surface electromyography," *(No Title)*, 1998.

[2]  C. J. De Luca, "Surface Electromyography: Detection and Recording. DelSys Incorporated (2002)," 2002, *Forschungsbericht*.

[3]  D. Farina, R. Merletti, and R. M. Enoka, "The extraction of neural strategies from the surface EMG," *J Appl Physiol*, vol. 96, no. 4, pp. 1486–1495, 2004.

[4]  P. Konrad, "The ABC of EMG: A practical introduction to kinesiological electromyography," 2005.

[5]  L. M. Vaca Benitez, M. Tabie, N. Will, S. Schmidt, M. Jordan, and E. A. Kirchner, "Exoskeleton Technology in Rehabilitation: Towards an EMG-Based Orthosis System for Upper Limb Neuromotor Rehabilitation," *Journal of Robotics*, vol. 2013, no. 1, p. 610589, 2013.

[6]  M. Jacquelin Perry, "Gait analysis: normal and pathological function," *New Jersey: SLACK*, 2010.

[7]  D. A. Neumann, "Kinesiology of the musculoskeletal system," *St. Louis: Mosby*, pp. 25–40, 2002.

[8]  P. Konrad, "The abc of emg," *A practical introduction to kinesiological electromyography*, vol. 1, no. 2005, pp. 30–35, 2005.

[9]  R. Merletti and P. J. Parker, *Electromyography: physiology, engineering, and non-invasive applications*, vol. 11. John Wiley & Sons, 2004.

[10] C. J. De Luca, "Surface electromyography: Detection and recording," *DelSys Incorporated*, vol. 10, no. 2, pp. 1–10, 2002.

[11] D. Xiong, D. Zhang, X. Zhao, and Y. Zhao, "Deep learning for EMG-based human-machine interaction: A review," *IEEE/CAA Journal of Automatica Sinica*, vol. 8, no. 3, pp. 512–533, 2021.

[12] S. M. Moghadam, T. Yeung, and J. Choisne, "A comparison of machine learning models' accuracy in predicting lower-limb joints' kinematics, kinetics, and muscle forces from wearable sensors," *Sci Rep*, vol. 13, no. 1, p. 5046, 2023.

[13] A. Gautam, M. Panwar, D. Biswas, and A. Acharyya, "MyoNet: A transfer-learning-based LRCN for lower limb movement recognition and knee joint angle prediction for remote monitoring of rehabilitation progress from sEMG," *IEEE J Transl Eng Health Med*, vol. 8, pp. 1–10, 2020.

[14] J. Han, H. Wang, and Y. Tian, "Multi Model Fusion Based Prediction of Human Joint Angles Using sEMG and Historical Angles," in *2023 42nd Chinese Control Conference (CCC)*, IEEE, 2023, pp. 3174–3179.

[15] J. Zhang *et al.*, "Physics-informed deep learning for musculoskeletal modeling: Predicting muscle forces and joint kinematics from surface EMG," *IEEE Transactions on Neural Systems and Rehabilitation Engineering*, vol. 31, pp. 484–493, 2022.

[16] A. Vijayvargiya, V. Gupta, R. Kumar, N. Dey, and J. M. R. S. Tavares, "A hybrid WD-EEMD sEMG feature extraction technique for lower limb activity recognition," *IEEE Sens J*, vol. 21, no. 18, pp. 20431–20439, 2021.

[17] A. Vijayvargiya, Khimraj, R. Kumar, and N. Dey, "Voting-based 1D CNN model for human lower limb activity recognition using sEMG signal," *Phys Eng Sci Med*, vol. 44, no. 4, pp. 1297–1309, 2021.

[19] M. W. Maciejewski, H. Z. Qui, I. Rujan, M. Mobli, and J. C. Hoch, "Nonuniform sampling and spectral aliasing," *Journal of Magnetic Resonance*, vol. 199, no. 1, pp. 88–93, 2009.

[20] J. Camargo, A. Ramanathan, W. Flanagan, and A. Young, "A comprehensive, open-source dataset of lower limb biomechanics in multiple conditions of stairs, ramps, and level-ground ambulation and transitions," *J Biomech*, vol. 119, p. 110320, 2021.

[21] O. F. A. Sanchez, J. L. R. Sotelo, M. H. Gonzales, and G. A. M. Hernandez, "Emg dataset in lower limb data set," *UCI machine learning repository*, vol. 2, 2014.

[22] F. H. Daryakenari, M. Mollahossein, A. Taheri, and G. R. Vossoughi, "Classification of lower limb electromyographical signals based on autoencoder deep neural network transfer learning," in *2022 10th RSI International Conference on Robotics and Mechatronics (ICRoM)*, IEEE, 2022, pp. 323–328.

[23] A. Vijayvargiya, C. Prakash, R. Kumar, S. Bansal, and J. M. R. S. Tavares, "Human knee abnormality detection from imbalanced sEMG data," *Biomed Signal Process Control*, vol. 66, p. 102406, 2021.

[24] A. Altıntaş and D. Yılmaz, "Classification of Knee Abnormality Using sEMG Signals with Boosting Ensemble Approaches," *Computer Science*, no. Special, pp. 48–52, 2021.

[25] X. Zhao, L. Wang, Y. Zhang, X. Han, M. Deveci, and M. Parmar, "A review of convolutional neural networks in computer vision," *Artif Intell Rev*, vol. 57, no. 4, p. 99, 2024.

[26] I. D. Mienye, T. G. Swart, and G. Obaido, "Recurrent neural networks: A comprehensive review of architectures, variants, and applications," *Information*, vol. 15, no. 9, p. 517, 2024.

[27] Z. Dai, D. Li, and S. Feng, "Attention Mechanism with Spatial-Temporal Joint Deep Learning Model for the Forecasting of Short-Term Passenger Flow Distribution at the Railway Station," *J Adv Transp*, vol. 2024, no. 1, p. 7985408, 2024.

[28] Mojtaba Hosseini, "https://github.com/MojtabaHosseiniie/."

[29] J. A. Raj, L. Qian, and Z. Ibrahim, "Fine-tuning--a Transfer Learning approach," *arXiv preprint arXiv:2411.03941*, 2024.